\documentclass{article}
\usepackage{spconf}

\usepackage{amsmath}
\usepackage{amsthm,amssymb,amsfonts,mathrsfs,mathtools,amsbsy}
\allowdisplaybreaks
\usepackage{bm}
\usepackage{algorithmic}
\usepackage{algorithm}
\usepackage{array}

\usepackage{siunitx}
\usepackage{cuted}
\usepackage{lipsum}

\usepackage{subcaption}
\usepackage{textcomp}
\usepackage{placeins}
\usepackage{url}
\usepackage{verbatim}
\usepackage{graphicx}
\usepackage[inline]{enumitem}
\usepackage{xspace}
\usepackage[table]{xcolor}
\usepackage[hidelinks]{hyperref}
\usepackage{tikz}
\usepackage{balance}

\colorlet{tableMulti}{red!20}
\colorlet{tableSingle}{cyan!30}

\newcommand{\IntegerP}{\mathbb{N}}

\newcommand{\Real}{\mathbb{R}}

\newcommand{\RealPP}{\mathbb{R}_{++}}

\newcommand\given{{\mathbin{}\mid\mathbin{}}}
\newcommand\vect[1]{\mathbf{#1}}
\newcommand\vectcal[1]{\mathbfcal{#1}}

\DeclarePairedDelimiter\norm{\lVert}{\rVert}
\DeclarePairedDelimiterX\innerp[2]{\langle}{\rangle}{#1
  \mathop{}\delimsize\vert\mathop{} #2}
\DeclarePairedDelimiter{\ceil}{\lceil}{\rceil}

\providecommand\given{} 
\newcommand\SetSymbol[1][]{
  \nonscript\,#1\vert \allowbreak \nonscript\,\mathopen{}}
\DeclarePairedDelimiterX\Set[1]{\lbrace}{\rbrace}%
{ \renewcommand\given{\SetSymbol[\delimsize]} #1 }

\DeclareMathAlphabet\mathbfcal{OMS}{cmsy}{b}{n}
\DeclareMathAlphabet\mathbfit{OML}{cmm}{b}{it}
\DeclareMathAlphabet\mathbfscr{OMS}{mdugm}{b}{n}

\usepackage[capitalize]{cleveref}

\crefname{thm}{Theorem}{Theorems}
\crefname{prop}{Proposition}{Propositions}
\crefname{assumption}{Assumption}{Assumptions}
\crefname{lemma}{Lemma}{Lemmata}
\crefname{definition}{Definition}{Definitions}
\crefname{example}{Example}{Examples}
\crefname{algo}{Algorithm}{Algorithms}
\crefname{fact}{Fact}{Facts}
\crefname{claim}{Claim}{Claims}
\crefname{appendix}{Appendix}{Appendices}
\crefname{coroll}{Corollary}{Corollaries}
\crefname{figure}{Figure}{Figures}
\crefname{section}{Section}{Sections}

\crefname{thmlisti}{Theorem}{Theorems}
\crefname{lemlisti}{Lemma}{Lemmata}
\crefname{proplisti}{Proposition}{Propositions}
\crefname{asslisti}{Assumption}{Assumptions}
\crefname{deflisti}{Definition}{Definitions}
\crefname{exlisti}{Example}{Examples}
\crefname{algolisti}{Algorithm}{Algorithms}
\crefname{factlisti}{Fact}{Facts}
\crefname{claimlisti}{Claim}{Claims}
\crefname{applisti}{Appendix}{Appendices}
\crefname{MyEnumSeci}{}{}
\crefname{MyEnumSubSeci}{}{}

\usepackage[%
backend = biber,%
maxbibnames = 10,%
minbibnames = 2,%
style = ieee,%
citestyle = numeric-comp,%
bibencoding = utf8,%
datamodel = software,%
defernumbers = true,%
sortcites=true,%
doi=true,%
isbn=false,%
url=false,%
eprint=false%
]{biblatex}

\makeatletter
\newcommand*{\ie}{%
  \@ifnextchar{,}%
  {i.e.}%
  {i.e.,\@\xspace}%
}
\newcommand*{\eg}{%
  \@ifnextchar{,}%
  {e.g.}%
  {e.g.,\@\xspace}%
}
\newcommand*{\cf}{%
  \@ifnextchar{,}%
  {cf.}%
  {cf.,\@\xspace}%
}
\makeatother

\usepackage{stackengine}
\stackMath
\newlength\matfield
\newlength\tmplength
\def\matscale{1.}
\newcommand\dimbox[4]{%
  \setlength\matfield{\matscale\baselineskip}%
  \setbox0=\hbox{\vphantom{X}\smash{#3}}%
  \setlength{\tmplength}{#1\matfield-\ht0-\dp0}%
  \fboxrule=1pt\fboxsep=-\fboxrule\relax%
  \fcolorbox{black}{#4}{\makebox[#2\matfield]{\addstackgap[.5\tmplength]{\box0}}}%
}
\newcommand\raiserows[2]{%
   \setlength\matfield{\matscale\baselineskip}%
   \raisebox{#1\matfield}{#2}%
}
\newcommand\matbox[6]{
  \stackunder{\dimbox{#1}{#2}{$#5$}{#6}}{\scriptstyle(#3\times #4)}%
}

\title{Imputation of time-varying edge flows in graphs by\\
  multilinear kernel regression and manifold learning}

\twoauthors{Duc Thien Nguyen\qquad Konstantinos Slavakis
  \thanks{This work was supported by JST SPRING, Japan Grant
    Number JPMJSP2106 and by the U.S. Air Force Office of Scientific Research, Agile Science of Test and Evaluation Program, under grants W911NF-23-S-0014 and W911NF-20-1-0283.}\vspace{-15pt}
}
{\small Tokyo Institute of Technology, Japan\\
  \footnotesize Department of Information and Communications
  Engineering\\[-5pt]
  \footnotesize Emails: \texttt{\{nguyen.t.au,
    slavakis.k.aa\}@m.titech.ac.jp} }
{Dimitris Pados \vspace{-16pt}}
{\small Florida Atlantic University, USA \\ [-5pt]
\footnotesize Center for Connected Autonomy and Artificial Intelligence \\ [-5pt]
\footnotesize Department of Electrical Engineering and Computer Science \\ [-5pt]
\footnotesize Email: \texttt{dpados@fau.edu}}

\begin{document}
\ninept
\sloppy
\maketitle

\begin{abstract}
  This paper extends the recently developed framework of
  multilinear kernel regression and imputation via manifold
  learning (MultiL-KRIM) to impute time-varying edge flows
  in a graph. MultiL-KRIM uses simplicial-complex arguments
  and Hodge Laplacians to incorporate the graph topology,
  and exploits manifold-learning arguments to identify
  latent geometries within features which are modeled as a
  point-cloud around a smooth manifold embedded in a
  reproducing kernel Hilbert space (RKHS). Following the
  concept of tangent spaces to smooth manifolds, linear
  approximating patches are used to add a
  collaborative-filtering flavor to the point-cloud
  approximations. Together with matrix factorizations,
  MultiL-KRIM effects dimensionality reduction, and enables
  efficient computations, without any training data or
  additional information. Numerical tests on real-network
  time-varying edge flows demonstrate noticeable
  improvements of MultiL-KRIM over several state-of-the-art
  schemes.
\end{abstract}




\section{Introduction}\label{sec:intro}

Graph signal processing~\cite{ortega2018graph,
  leus2023graph,multilkrim} plays a pivotal role in modern
data analytics, because signals are often associated with
entities that have physical or intangible inter-connections,
which are usually modeled by graphs. Due to reasons such as
user privacy, sensor fault, or resources conservation,
signal observation/measurement is often incomplete (missing
data), causing bias and errors in subsequent stages of
learning~\cite{chen2015signal, chen2016signal}. Much has
been done to address the missing-data problem when observed
signals are associated with the nodes of a graph,
\eg,~\cite{qiu2017time, romero2017kernel, correa2024gegen,
  multilkrim}. Nevertheless, many systems observe
\textit{signals over edges (edge flows),} with the relative
research gaining attention only
recently~\cite{battiston2020networks,
  barbarossa2020topological, schaub2021signal}.

Node-signal imputation techniques may not be successfully
applied to the edge-flow imputation problem. For example,
given a graph with observed signals over edges, one can
consider its line-graph counterpart~\cite{evans2009line}
whose nodes are edges of the original graph. In this way,
the original problem is reduced to a node-signal imputation
one, often solved by imposing spatial smoothness via a graph
Laplacian matrix. However, this approach has been shown to
produce subpar solutions as the smoothness assumption no
longer holds for edge flows~\cite{schaub2018flow,
  jia2019graph}. Instead, it is more effective to assume
that edge flows are almost \textit{divergence-free,} \ie,
the inbound at a node is approximately equal to the
outbound, and \textit{curl-free,} \ie, the total flow along
a ``triangle'' is close to zero~\cite{schaub2018flow,
  jia2019graph, barbarossa2020topological, schaub2021signal,
  yang2022trend}. Such assumptions are efficiently modeled
by the theory of simplicial complexes and Hodge Laplacians
on graphs~\cite{lim2020hodge}. These theories bring forth
also simplicial convolutional filters, defined as matrix
polynomials of Hodge Laplacians, which model the multi-hop
shift of signals across simplicial
complexes~\cite{isufi2022convolutional, yang2022simplicial}.

While such theories are capable of spatial (graph topology)
description, vector autoregression (VAR) is commonly used to
model dependencies along time whenever edge flows are time
varying~\cite{money2022online}. However, VAR is negligent of
the underlying graph topology. Recent efforts fuse
simplicial convolutional filters into VAR, coined simplicial
VAR (S-VAR), to take the graph topology into account, as
well as to reduce the number of coefficients to be learned
(dimensionality reduction)~\cite{krishnan2023simplicial,
  krishnan2024simplicial,
  money2024evolution}. Notwithstanding, (S-)VAR relies by
definition on past observations to predict/regress future
ones, which may be problematic under the presence of missing
data. Besides, VAR-based models are linear, and may thus
fail to capture intricate data and feature dependencies. To
address this issue, there have been studies incorporating
simplicial complexes into (non-linear) neural
networks~\cite{ebli2020simplicial,
  roddenberry2021principled, yang2022simpnn,
  wu2023simplicial}, but typically, such models require
large amounts of training data and intricate concatenation
of non-linear layers. Further, it seems that the existing
literature on simplicial-complex neural networks has not
considered yet the problem of imputation of time-varying
edge flows.

This paper extends the recently developed method of
multilinear kernel regression and imputation via manifold
learning (MultiL-KRIM)~\cite{multilkrim}, and applies it to
the problem of edge-flow imputation. MultiL-KRIM has been
already applied to node-signal
imputation~\cite{multilkrim}. Unlike (S-)VAR, where data
depend directly on past observations, MultiL-KRIM assumes
that missing entries can be estimated by a set of landmark
points, extracted from measurements and located around a
smooth manifold, embedded in an ambient reproducing kernel
Hilbert space (RKHS). As such, functional approximation is
enabled by the RKHS, effecting thus nonlinear data modeling,
which is a lacking attribute in VAR models.  Furthermore,
while S-VAR applies simplicial convolutional filters for
dimensionality reduction, MultiL-KRIM achieves this by
matrix factorizations. Hodge Laplacians are incorporated in
MultiL-KRIM's inverse problem to take account of the graph
topology. In addition, MultiL-KRIM's manifold and
tangent-space arguments model ``locality,'' captured in
S-VAR by the number of hops in the simplicial convolutional
filter. Unlike neural networks, MultiL-KRIM needs no
training data and offers a more explainable approach through
intuitive geometric arguments. Numerical tests on real
networks show that MultiL-KRIM outperforms state-of-the-art
methods.

\section{Preliminaries}\label{sec:prem}

A graph is denoted by $G = (\mathcal{V}, \mathcal{E})$,
where $\mathcal{V}$ represents the set of nodes, and
$\mathcal{E} \subseteq \mathcal{V} \times \mathcal{V}$ is
the set of edges. A $k$-simplex
$\mathcal{S}^k \subset \mathcal{V}$ comprises $k+1$ distinct
elements of $\mathcal{V}$. For example, a 0-simplex is a
node, an 1-simplex is an edge, and a 2-simplex is a
triangle. A simplicial complex (SC) $\mathcal{X}$ is a
finite collection of simplices with the inclusion property:
for any simplex $\mathcal{S}^k \in \mathcal{X}$, if
$\mathcal{S}^{k-1} \subset \mathcal{S}^{k}$, then
$\mathcal{S}^{k-1} \in \mathcal{X}$. An SC of order $K$,
denoted as $\mathcal{X}^K$, contains at least one
$K$-simplex~\cite{lim2020hodge, barbarossa2020topological}.

Hodge Laplacians~\cite{lim2020hodge} describe adjacencies in
an SC. Specifically, in an SC $\mathcal{X}^K$, the incidence
matrix $\vect{B}_k\in \Real^{N_{k-1}\times N_k}, k\geq 1$
captures the adjacencies between $(k-1)$- and $k$-simplices,
where $N_k$ is the number of $k$-simplicies in
$\mathcal{X}$.  For example, $\vect{B}_1$ is the
node-to-edge incidence matrix and $\vect{B}_2$ is the
edge-to-triangle one.  Incidence matrices satisfy the
boundary condition
$\vect{B}_k\vect{B}_{k+1}=\vect{0}, \forall k \geq 1$. Hodge
Laplacians are defined as
\begin{align}
  \vect{L}_k \coloneqq \begin{cases}
    \vect{B}_{k+1} \vect{B}_{k+1}^\intercal\,,
    & \text{if } k = 0 \,,
    \\
    \vect{B}_{k}^\intercal\vect{B}_{k} + \vect{B}_{k+1}
    \vect{B}_{k+1}^\intercal \,,
    & \text{if } 1\leq k \leq K-1
      \,, \\
    \vect{B}_{k}^\intercal\vect{B}_{k} \,,
    & \text{if } k=K \,,
      \label{eq:hodge}
  \end{cases}
\end{align}
where $\vect{L}_0$ is the well-known graph Laplacian matrix,
which describes node-adjacencies via shared edges. Hodge
Laplacian $\vect{L}_1$ defines the adjacencies between edges
via shared nodes by the lower-Laplacian
$\vect{L}_{1,l} \coloneqq
\vect{B}_{1}^\intercal\vect{B}_{1}$, and via shared
triangles by the upper-Laplacian
$\vect{L}_{1,u}\coloneqq
\vect{B}_{2}\vect{B}_{2}^\intercal$. Likewise, $\vect{L}_2$
denotes the connection between triangles through common
edges.  As this paper concerns signals over edges,
$\vect{L}_1$ will be the main focus.

``Simplicial signals'' are abstracted as functions which map
a $k$-simplex to a real number. In case of time-varying
signals, an ``edge-flow signal'' at time
$t \in\{ 1, 2, \ldots, T \}$ is denoted by
$\vect{y}_t = [y_{1t}, y_{2t}, \ldots, y_{N_1
  t}]^\intercal\in \Real^{N_1}$. These vectors comprise the
data matrix
$\vect{Y} = [\vect{y}_1, \vect{y}_2, \ldots, \vect{y}_T] \in
\Real^{N_1\times T}$.  To account for missing entries, let
the index set of observed entries
$\Omega \coloneqq \Set{ (i,t) \in \Set{1, \ldots, N_1}
  \times \Set{1, \ldots, T} \given y_{it} \neq +\infty }$,
where $+\infty$ denotes a missing entry, and define the
linear \textit{sampling mapping}\/
$\mathscr{S}_{\Omega} \colon (\Real \cup \{ +\infty \})^{N_1
  \times T} \to \Real^{N_1 \times T} \colon \vect{Y} \mapsto
\mathscr{S}_{\Omega}(\vect{Y})$, which operates entry-wisely
as follows:
$[\mathscr{S}_{\Omega}(\vect{Y})]_{it} \coloneqq
[\vect{Y}]_{it}$, if $(i,t) \in \Omega$, while
$[\mathscr{S}_{\Omega}(\vect{Y})]_{it} \coloneqq 0$, if
$(i,t) \notin \Omega$.

An edge-flow imputation framework solves the following
inverse problem
\begin{align}
  \min\nolimits_{ \vect{X}\in \Real^{N_1 \times T} }
  {} & {} \mathcal{L}( \vect{X} ) + \mathcal{R}(\vect{X}) {}
       \notag\\
  \text{s.to}\ {}
     & {} \mathscr{S}_\Omega(\vect{Y}) = \mathscr{S}_\Omega
       (\vect{X})\ \text{and other
       constraints,} \label{recovery.generic.generic}
\end{align}
where $\mathcal{L}(\cdot)$ is the data-fit loss and
$\mathcal{R}(\cdot)$ is the regularizer for structural
priors, while constraint
$\mathscr{S}_\Omega(\vect{Y}) = \mathscr{S}_\Omega
(\vect{X})$ preserves the consistency of the observed
entries of $\vect{Y}$. For example,
FlowSSL~\cite{jia2019graph} sets
$\mathcal{L}(\vect{X}) \coloneqq 0$ and
$\mathcal{R}(\vect{X}) \coloneqq ({\lambda_l}/{2})
\norm{\vect{B}_1 \vect{X}}_{\textnormal{F}}^2 +
({\lambda_u}/{2}) \norm{\vect{B}_2^\intercal
  \vect{X}}_{\textnormal{F}}^2$, which imposes the
conservation of the flows at the nodes and cyclic flows
along edges of triangles. More recent studies incorporate
the Hodge Laplacians into VAR, a widely used tool for
time-varying signals modeling.  In particular,
S-VAR~\cite{krishnan2023simplicial, krishnan2024simplicial,
  money2024evolution} approximates each snapshot
$\vect{y}_t\approx \sum\nolimits_{p=1}^P
\vect{H}_p(\vect{L}_1) \vect{y}_{t-p}$, where
$\Set{\vect{H}_p(\vect{L}_1)}_{p=1}^P$ are simplicial
convolutional filters~\cite{yang2022simplicial}, whose role
is to capture the multi-hop dependencies between edges.  As
a result, S-VAR sets
$\mathcal{L}(\vect{X}) \coloneqq \sum \nolimits_{t=1}^T
\norm{\vect{x}_t-\vect{F}_t \bm{\beta}_t}_2^2$, where
$\Set{\bm{\beta}_t}_{t=1}^T$ are learnable parameters, and
$\Set{\vect{F}_t}_{t=1}^T$ are ``shifted signals'' of
$\Set{\vect{y}_t}_{t=1}^T$.


\section{Proposed method} \label{sec:modeling}

\subsection{A short recap of MultiL-KRIM}

\begin{figure}[t!]
  \centering
  \includegraphics[width =
  .7\columnwidth]{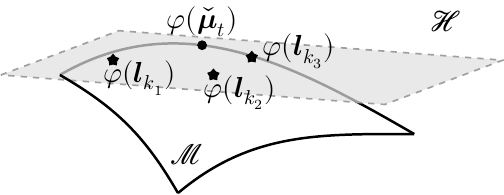}
  \caption{A ``collaborative-filtering'' approach: Points
    $\Set{ \varphi (\mathbfit{l}_{k_j}) }_{j=1}^3$, which
    lie into or close to the unknown-to-the-user manifold
    $\mathscr{M} \subset \mathscr{H}$, collaborate affinely
    to approximate $\varphi (\check{\bm{\mu}}_{t})$. All
    affine combinations of
    $\Set{ \varphi (\mathbfit{l}_{k_j})}_{j=1}^3$ define the
    approximating ``linear patch'' (gray-colored plane),
    which mimics the concept of a tangent space to
    $\mathscr{M}$~\cite{RobbinSalamon:22}.}
  \label{fig:manifold.kernel.space}
\end{figure}

MultiL-KRIM~\cite{multilkrim} is a generic
data-imputation-by-regression framework which combines
matrix factorization, kernel methods and manifold
learning. The approach can be shortly described as follows.

Assuming that data $\mathscr{S}_\Omega(\vect{Y})$ are
faithful, a subset of $\mathscr{S}_\Omega(\vect{Y})$, called
\textit{navigator/pilot} data and denoted as
$\check{\vect{Y}}_{\text{f}} \coloneqq [
\check{\vect{y}}_1^{\text{f}}, \ldots,
\check{\vect{y}}_{N_\text{nav}}^{\text{f}}] \in \Real^{\nu
  \times N_\text{nav}}$, are selected by the user; for
example, all of $\mathscr{S}_\Omega(\vect{Y})$ can be
selected. As the cardinality of the point-cloud
$\{ \check{\vect{y}}^{\text{f}}_t \}_{t=1}^{N_{\text{nav}}}$
grows for large datasets, a subset
$\{ \mathbfit{l}_k \}_{k=1}^{N_{\mathit{l}}}$, coined
\textit{landmark/representative}\/ points, with
$N_{\mathit{l}} \leq N_{\text{nav}}$, is selected from
$\{ \check{\vect{y}}^{\text{f}}_t
\}_{t=1}^{N_{\text{nav}}}$. Several ways to form navigator
data and landmark points have been
suggested~\cite{multilkrim}. Let
$\check{\vect{L}} \coloneqq [ \mathbfit{l}_1,
\mathbfit{l}_2, \ldots, \mathbfit{l}_{N_{\mathit{l}}}] \in
\Real^{\nu \times N_{\mathit{l}}}$.  To facilitate nonlinear
approximation, a feature map $\varphi(\cdot)$ takes vector
$\mathbfit{l}_k$ to $\varphi(\mathbfit{l}_k)$ in an RKHS
$\mathscr{H}$, equipped with a reproducing kernel
$\kappa(\cdot, \cdot): \Real^{\nu}\times \Real^{\nu} \to
\Real$, with well-documented merits in approximation
theory~\cite{aronszajn1950theory}.  To this end,
$\varphi(\mathbfit{l}) \coloneqq \kappa(\mathbfit{l}, \cdot
) \in \mathscr{H}$, $\forall \mathbfit{l}\in \Real^{\nu}$.
For convenience, let
$\bm{\Phi}( \check{\vect{L}} ) \coloneqq [
\varphi(\mathbfit{l}_1), \ldots, \varphi(
\mathbfit{l}_{N_{\mathit{l}}}) ]$. Consequently,
$\vect{K} \coloneqq \bm{\Phi}( \check{\vect{L}} )^\intercal
\bm{\Phi}( \check{\vect{L}} )$ is an
$N_{\mathit{l}} \times N_{\mathit{l}}$ kernel matrix whose
$(k, k^{\prime})$th entry is equal to
$\innerp{\varphi(\mathbfit{l}_k)}
{\varphi(\mathbfit{l}_{k^{\prime}})}_{\mathscr{H}} =
\kappa(\mathbfit{l}_k, \mathbfit{l}_{k^{\prime}})$, where
$\intercal$ stands for vector/matrix transposition, and
$\innerp{\cdot}{\cdot}_{\mathscr{H}}$ for the inner product
of the RKHS $\mathscr{H}$.

The flow $x_{it}$ through the edge $i$ at time $t$ is
approximated as
\begin{alignat}{2}
  [\vect{X}]_{it} = x_{it}
  & {} \approx {}
  && f_i( \check{\bm{\mu}}_{t}
     ) = \innerp{f_i}{ \varphi(
     \check{\bm{\mu}}_{t})}_{\mathscr{H}}
     \,, \label{xij.reproducing.property}
\end{alignat}
where $f_i(\cdot): \Real^{\nu} \to \Real$ is an unknown
non-linear function in the RKHS functional space
$\mathscr{H}$, $\check{\bm{\mu}}_{t}$ is an unknown vector
in $\Real^{\nu}$, and the latter part
of~\eqref{xij.reproducing.property} is because of the
reproducing property of
$\mathscr{H}$~\cite{aronszajn1950theory}.  Motivated by the
representer theorem, $f_i$ is assumed to belong to the
linear span of
$\Set{ \varphi( \mathbfit{l}_k) }_{k=1}^{ N_{\mathit{l}} }$,
\ie there exists
$\vect{u}_i \coloneqq [ u_{i1}, \ldots,
u_{iN_{\mathit{l}}}]^{\intercal} \in \Real^{N_{\mathit{l}}}$
s.t.\
$f_i = \sum_{k=1}^{N_{\mathit{l}}} u_{ik} \varphi(
\mathbfit{l}_k ) = \bm{\Phi}( \vect{L} ) \vect{u}_i$.
Regarding $\varphi( \check{\bm{\mu}}_{t})$, the concept of
tangent spaces to smooth manifolds~\cite{RobbinSalamon:22}
offers also a ``collaborative-filtering'' flavor to the
design: it is assumed that $\varphi( \check{\bm{\mu}}_{t})$
lies into or close to $\mathscr{M}$ and is approximated by
only a few members of
$\Set{ \varphi( \mathbfit{l}_k) }_{k=1}^{ N_{\mathit{l}} }$
which collaborate \textit{affinely,} \ie, there exists a
sparse vector
$\vect{v}_t \in \Real^{N_{\mathit{l}}}$ s.t.\
$\varphi( \check{\bm{\mu}}_{t}) = \bm{\Phi}(\vect{L})
\vect{v}_t$, under the affine constraint
$\vect{1}_{N_{\mathit{l}}}^{{\intercal}} \vect{v}_t = 1$,
where $\vect{1}_{N_{\mathit{l}}}$ is the
$N_{\mathit{l}} \times 1$ all-one vector; \cf
\cref{fig:manifold.kernel.space}.

So far, the entry-wise estimation
$[\vect{X}]_{it} \approx f_i( \check{\bm{\mu}}_{t} ) =
\innerp{f_i}{ \varphi( \check{\bm{\mu}}_{t})}_{\mathscr{H}}
= \innerp{\bm{\Phi}( \vect{L} ) \vect{u}_i}
{\bm{\Phi}(\vect{L}) \vect{v}_t}_{\mathscr{H}} =
\vect{u}_i^{\intercal} \vect{K} \vect{v}_t$ . To offer
compact notations, if
$\vect{U} \coloneqq [ \vect{u}_1, \ldots, \vect{u}_{N_1}
]^{{\intercal}} \in \Real^{ N_1 \times N_{\mathit{l}}}$ and
$\vect{V} \coloneqq [ \vect{v}_1, \ldots, \vect{v}_{T} ] \in
\Real^{ N_{\mathit{l}} \times T }$, then data are modeled as
\begin{align}
    \vect{X} \approx \vect{U} \vect{K} \vect{V}
  \,, \label{eq:ukv}
\end{align}
where $\vect{V}$ is a sparse matrix satisfying
$\vect{1}_{N_\mathit{l}}^{\intercal} \vect{V} =
\vect{1}_{T}^{\intercal}$. To ease the excessive cost
caused by large $N_1$ and $N_\mathit{l}$, \cite{multilkrim}
employs multilinear factorization
$\vect{U} = \vect{U}_1 \vect{U}_2 \cdots \vect{U}_Q$, which
has been shown to be highly efficient, especially for
high-dimensional dynamic imaging data~\cite{multilkrim}.

\subsection{Extended MultiL-KRIM for edge-flow imputation}

In large networks with long time series, sizes $N_1, T,$
and $N_{\mathit{l}}$ can cause massive burden. Unlike
S-VAR~\cite{krishnan2023simplicial, krishnan2024simplicial,
  money2024evolution}, which reduces the model size of VAR by
simplicial convolutional filters, and \cite{multilkrim},
which factors only $\vect{U}$ in~\eqref{eq:ukv}, this paper
suggests the following factorization
\renewcommand\matscale{.5}
\begin{align}
  \vect{X}
  & \approx \vect{U}_1
 \vect{U}_2 \vect{K}
    \vect{V}_1 \vect{V}_2
    \,, \label{eq:multi.kernel.nodimred} \\ 
  \matbox{9}{4}{N_1}{T}{\vect{X}}{tableSingle}
  & \approx
    {\matbox{9}{3}{N_1}{d}{\vect{U}_1}{tableSingle}} 
    \raiserows{3.25}
    {\matbox{2.5}{4}{d}
    {N_{\mathit{l}}}{\vect{U}_2}{tableSingle}}
    \raiserows{2.5}
    {\matbox{4}{4}{N_{\mathit{l}}}
    {N_{\mathit{l}}}{\vect{K}}{white}}
    \raiserows{2.5}
    {\matbox{4}{2}{N_{\mathit{l}}}
    {r}{\vect{V}_1}{tableSingle}} 
    \raiserows{3.5}
    {\matbox{2}{4}{r}
    {T}{\vect{V}_2}{tableSingle}} \notag
\end{align}
where the inner matrix dimensions $d, r \ll N_\mathit{l}$
are user-defined.  Based on findings in~\cite{multilkrim},
factorizing $\vect{U}$ further, by more than two factors,
does not notably improve the accuracy, so
$\vect{U} = \vect{U}_1 \vect{U}_2$ and
$\vect{V} = \vect{V}_1 \vect{V}_2$ are used here. It can be
easily checked that the affine constraint of $\vect{V}$
in~\eqref{eq:ukv} still holds after enforcing
$\vect{1}_{N_\mathit{l}}^{\intercal} \vect{V}_1 =
\vect{1}_{r}^{\intercal}$ and
$\vect{1}_{r}^{\intercal} \vect{V}_2 =
\vect{1}_{T}^{\intercal}$.

Ultimately, following the common assumption that edge flows
are approximately \textit{divergence-free} and
\textit{curl-free} (\cf \cref{sec:prem}), the extended
MultiL-KRIM inverse problem for edge-flow imputation is
formed as
\begin{subequations}\label{eq:graph.task.general}
  \begin{align}
    \min_{ (\vect{X}, \vect{U}_1, \vect{U}_2, \vect{V}_1,
    \vect{V}_2) }
    {}\ & {} \tfrac{1}{2} \norm{ \vect{X} - \vect{U}_1
          \vect{U}_2 \vect{K}
          \vect{V}_1 \vect{V}_2 }^2_{\textnormal{F}}  \notag \\
        & + \tfrac{\lambda_l}{2} \norm{\vect{B}_1
          \vect{X}}_{\textnormal{F}}^2
        + \tfrac{\lambda_u}{2} \norm{\vect{B}_2^\intercal
          \vect{X}}_{\textnormal{F}}^2 \notag \\
          & + \tfrac{\lambda_2}{2}
          \sum_{q=1}^2 \norm{\vect{U}_q}_{\textnormal{F}}^2 + {
          \lambda_1 \sum_{p=1}^2 \norm{\vect{V}_p}_1 }
           \label{eq:graph.task.loss} \\
    \text{s.to} {}\
        & \mathscr{S}_{\Omega}(\vect{Y}) = \mathscr{S}_{\Omega}
          (\vect{X})\,, \label{graph.task.consistency} \\
        & \vect{1}_{N\mathit{l}}^{\intercal} \vect{V}_1
          = \vect{1}_{r}^{\intercal} \,, \notag \\
        & \vect{1}_{r}^{\intercal} \vect{V}_2
          = \vect{1}_{T}^{\intercal}
          \,. \label{graph.task.right}
  \end{align}
\end{subequations}
The loss function is non-convex, and to guarantee
convergence to a critical point, the parallel
successive-convex-approximation (SCA) framework
of~\cite{facchinei2015parallel} is utilized. The algorithm
is summarized in \cref{alg:multil.general}, where the
following tuple of estimates
$\forall n\in\IntegerP, \forall k\in\Set{0,1}$,
\begin{align}
    \hat{\mathbfcal{O}}^{(n+k/2)} \coloneqq (
      {} & {} \hat{\vect{X}}^{(n+k/2)},
           \hat{\vect{U}}_1^{(n+k/2)}, \notag \\
         & \hat{\vect{U}}_2^{(n+k/2)},
           \hat{\vect{V}}_1^{(n+k/2)}, \hat{\vect{V}}_2^{(n+k/2)}
           )\,, \label{Oh.tuple.tvgs}
\end{align}
is recursively updated via the following convex sub-tasks
which can be solved in parallel at every iteration $n$:
\begin{subequations}\label{eq:graph.subtasks}
  \begin{alignat}{3}
    && \hat{\vect{X}}^{(n + 1/2) }
    && \coloneqq \arg\min_{ {\vect{X}} } {}
    & {}\ \tfrac{1}{2} \norm{{\vect{X}} -
      \hat{\vect{U}}_1^{(n)}
      \hat{\vect{U}}_2^{(n)} \vect{K}
      \hat{\vect{V}}_1^{(n)} \hat{\vect{V}}_2^{(n)}
      }_{\textnormal{F}}^2
      \notag \\
    &&&&& {}\ + \tfrac{\lambda_l}{2} \norm{\vect{B}_1
          \vect{X}}_{\textnormal{F}}^2
        + \tfrac{\lambda_u}{2} \norm{\vect{B}_2^\intercal
          \vect{X}}_{\textnormal{F}}^2 \notag \\
    &&&&& {}\ + \tfrac{\tau_X}{2} \norm{
          \vect{X} - \hat{\vect{X}}^{(n)
          }}_{\textnormal{F}}^2 \notag \\
    &&&& \text{s.to} {} & {}\ \mathscr{S}_{\Omega}(\vect{Y})
                          = \mathscr{S}_{\Omega} (
                          {\vect{X}} )
                          \,, \label{eq:graph.min.X} \\
    && \hat{\vect{U}}_1^{(n + 1/2)} {}
    && {} \coloneqq \arg\min_{\vect{U}_1} {}
    & {}\ \tfrac{1}{2} \norm{ \hat{\vect{X}}^{(n)} -
      \vect{U}_1 \hat{\vect{U}}_2^{(n)} \vect{K}
      \hat{\vect{V}}_1^{(n)} \hat{\vect{V}}_2^{(n)}
      }_{\textnormal{F}}^2 \notag \\
    &&&& {} & {}\ + \tfrac{\lambda_2}{2}
              \norm{\vect{U}_1}_{\textnormal{F}}^2 +
              \tfrac{\tau_U}{2} \norm{ \vect{U}_1 -
              \hat{\vect{U}}_1^{(n)}
              }_{\textnormal{F}}^2
                          \,, \label{eq:graph.min.U1} \\
    && \hat{\vect{U}}_2^{(n + 1/2)} {}
    && {} \coloneqq \arg\min_{\vect{U}_2} {}
    & {}\ \tfrac{1}{2} \norm{ \hat{\vect{X}}^{(n)} -
      \hat{\vect{U}}_1^{(n)}
      \vect{U}_2 \vect{K}
      \hat{\vect{V}}_1^{(n)} \hat{\vect{V}}_2^{(n)}
      }_{\textnormal{F}}^2 \notag \\
    &&&& {} & {}\ + \tfrac{\lambda_2}{2}
              \norm{\vect{U}_2}_{\textnormal{F}}^2 +
              \tfrac{\tau_U}{2} \norm{ \vect{U}_2 -
              \hat{\vect{U}}_2^{(n)}
              }_{\textnormal{F}}^2
                          \,, \label{eq:graph.min.U2} \\
    && \hat{\vect{V}}_1^{(n + 1/2)}
    && \coloneqq \arg \min_{\vect{V}_1} {}
    & {}\ \tfrac{1}{2} \norm{ \hat{\vect{X}}^{(n)}-
      \hat{\vect{U}}_1^{(n)}
      \hat{\vect{U}}_2^{(n)} \vect{K}
      \vect{V}_1 \vect{V}_2^{(n)}
       }_{\textnormal{F}}^2 \notag \\
    &&&&& {}\ + \lambda_1 \norm{\vect{V}_1}_1 +
          \tfrac{\tau_V}{2} \norm{ \vect{V}_1 -
          \hat{\vect{V}}_1^{(n)} }_{\textnormal{F}}^2 \notag \\
    &&&& \text{s.to} {}
    & {}\ \vect{1}_{N_\mathit{l}}^{{\intercal}} \vect{V}_1
      = \vect{1}_{r}^{{\intercal}}
      \,, \label{eq:graph.min.V1} \\
      && \hat{\vect{V}}_2^{(n + 1/2)}
    && \coloneqq \arg \min_{\vect{V}_2} {}
    & {}\ \tfrac{1}{2} \norm{ \hat{\vect{X}}^{(n)}-
      \hat{\vect{U}}_1^{(n)}
      \hat{\vect{U}}_2^{(n)} \vect{K}
      \vect{V}_1^{(n)} \vect{V}_2
       }_{\textnormal{F}}^2 \notag \\
    &&&&& {}\ + \lambda_1 \norm{\vect{V}_2}_1 +
          \tfrac{\tau_V}{2} \norm{ \vect{V}_2 -
          \hat{\vect{V}}_2^{(n)} }_{\textnormal{F}}^2 \notag \\
    &&&& \text{s.to} {}
    & {}\ \vect{1}_{r}^{{\intercal}} \vect{V}_2
      = \vect{1}_{T}^{{\intercal}}
      \,, \label{eq:graph.min.V2}
  \end{alignat}
\end{subequations}
where the user-defined
$\tau_X, \tau_U, \tau_V, \lambda_1, \lambda_2, \lambda_l,
\lambda_u \in \RealPP$.  Sub-tasks \eqref{eq:graph.min.V1}
and \eqref{eq:graph.min.V2} are composite convex
minimization tasks under affine constraints, and can be thus
solved iteratively by~\cite{slavakis2018fejer},
while~\eqref{eq:graph.min.X}, \eqref{eq:graph.min.U1},
and~\eqref{eq:graph.min.U2} have closed-form solutions.

\subsection{Computational complexity}

The number of unknown parameters
in~\eqref{eq:multi.kernel.nodimred} is
$(N_1+N_\mathit{l})d + (N_\mathit{l}+T)r$. Per iteration $n$
in~\cref{alg:multil.general}, the computational complexity
of sub-task~\eqref{eq:graph.min.U1} is
$\mathcal{O}(N_1 d^2 + d^3)$, and that of
\eqref{eq:graph.min.U2} is
$\mathcal{O}(d^3 + N_\mathit{l}^3)$. Meanwhile, the
complexity of sub-task~\eqref{eq:graph.min.V1} is
$\mathcal{O}(K_1 N_\mathit{l}^2 d)$, and that of
\eqref{eq:graph.min.V2} is $\mathcal{O}(K_2 d^2 T)$, where
$K_1$ and $K_2$ are the numbers of iterations that
\cite{slavakis2018fejer} takes to solve
\eqref{eq:graph.min.V1} and~\eqref{eq:graph.min.V2},
respectively.

\begin{algorithm}[!t]
  \caption{Solving MultiL-KRIM's inverse
    problem}\label{alg:multil.general}
  \begin{algorithmic}[1]

  \REQUIRE

    \ENSURE Limit point $\hat{\vectcal{O}}^{(*)}$ of sequence
    $(\hat{\vectcal{O}}^{(n)})_{n\in\IntegerP}$.

    \STATE Fix $\hat{\mathbfcal{O}}^{(0)}$, $\gamma_0\in
    (0,1]$, and $\zeta\in (0,1)$.

    \WHILE{$n\geq 0$} \label{alg.step:resume.k}

    \STATE {Available are
      $\hat{\vectcal{O}}^{(n)}$ \eqref{Oh.tuple.tvgs}.}

    \STATE {$\gamma_{n+1} \coloneqq \gamma_n (1 - \zeta
      \gamma_n)$.}

    \STATE Solve in parallel the convex sub-tasks
    \eqref{eq:graph.subtasks}.

    \STATE
    {$\hat{\vectcal{O}}^{(n+1)} \coloneqq \gamma_{n+1}
      \hat{\vectcal{O}}^{(n+1/2)} + (1-\gamma_{n+1})
      \hat{\vectcal{O}}^{(n)}$.}

    \STATE {Set $n \leftarrow n+1$ and go to
      step~\ref{alg.step:resume.k}.}

    \ENDWHILE
  \end{algorithmic}
\end{algorithm}


\section{Numerical Tests}\label{sec:numerical}

MultiL-KRIM is tested on the traffic flows in Sioux Falls
transportation network~\cite{rossman1994modeling} and on the
water flows in the Cherry Hills water
network~\cite{leblanc1975algorithm}. MultiL-KRIM is compared
against the state-of-the-art edge-flow imputation methods
FlowSSL~\cite{jia2019graph} and S-VAR
\cite{krishnan2023simplicial, krishnan2024simplicial,
  money2024evolution}. As a baseline for
matrix-factorization techniques, the multi-layer matrix
factorization (MMF)~\cite{cichocki2007multilayer} is
implemented. Here, MMF~\cite{cichocki2007multilayer} solves
the same inverse problem~\eqref{eq:graph.task.general} as
MultiL-KRIM, but with the kernel matrix
$\vect{K} = \vect{I}_{N_\mathit{l}}$
in~\eqref{eq:multi.kernel.nodimred}. In other words,
MMF~\cite{cichocki2007multilayer} is a \textit{blind}\/
matrix-factorization method, where no data geometry/patterns
are explored and exploited, as in
\cref{fig:manifold.kernel.space}, and no functional
approximation via RKHSs is employed.

The evaluation metric is the
mean absolute error (MAE) (lower is better), defined as
$\text{MAE} \coloneqq {\norm{\vect{X} - \vect{Y}}_1} /
({N_1\cdot T})$, where $\vect{Y}$ is the fully-sampled data,
$\vect{X}$ gathers all reconstructed flows, and
$\norm{\cdot}_1$ is the $\ell_1$-norm.  All methods are
finely tuned to achieve their lowest MAE. Reported metric
values are mean values of \num{30} independent runs. Source
codes of FlowSSL~\cite{jia2019graph} and
S-VAR~\cite{krishnan2023simplicial, krishnan2024simplicial,
  money2024evolution} were made publicly available by the
authors. Source code for MultiL-KRIM and MMF was written in
Julia~\cite{bezanson2017julia}. All tests were run on an
8-core Intel(R) i7-11700 2.50GHz CPU with 32GB RAM.

\begin{figure}[t!]
  \centering
  \subfloat[Cherry Hills water network \label{fig:plot.water}]
  {\includegraphics[width =
    .7\columnwidth]{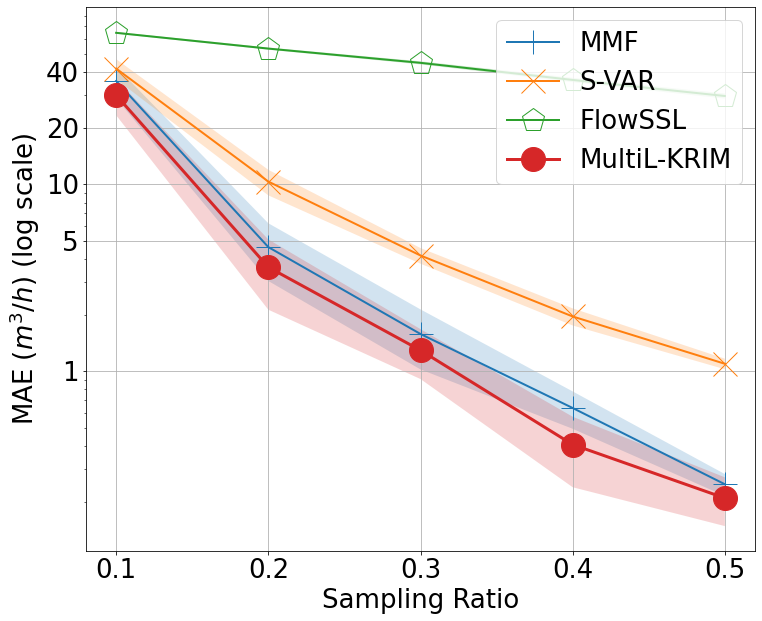}} \\
    \subfloat[Sioux Falls traffic
    network \label{fig:plot.traffic}]
  {\includegraphics[width = .7\columnwidth]{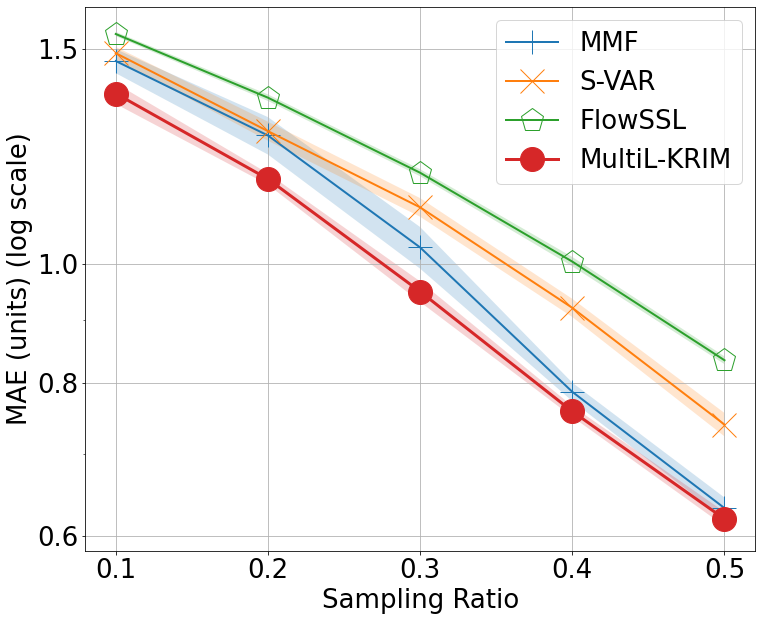}}
  \centering
  \caption{Mean MAE value curves in log scale (the lower the
    better) w.r.t.\ the groud truth signals vs.\ sampling
    ratios. The shaded areas indicate a range of one
    standard deviation above and below the mean.
  } \label{fig:mae}
\end{figure}

With sampling ratio $s \in \Set{0.1, 0.2, 0.3, 0.4, 0.5}$,
signals of $\ceil{N_1 \cdot s}$ edges are sampled per time
instant $t$, where $\ceil{\cdot}$ is the ceiling function.
This sampling pattern suggests that the number of
observations is consistent along time. Navigator data
$\check{\vect{Y}}_{\textnormal{nav}}$ are formed by the
snapshots, that is, columns of
$\mathscr{S}_\Omega(\vect{Y})$ after removing all the
unobserved entries. Landmark points are selected by the
greedy max-min-distance strategy \cite{de2004sparse}, based
on Euclidean distances among navigator data. Hyperparameters
$\tau_X = \tau_U = \tau_V =2$ in~\eqref{eq:graph.subtasks}
to ensure stability of the algorithm. Meanwhile,
regularization hyperparameters
$\lambda_1, \lambda_2, \lambda_l$, and $\lambda_u$
in~\eqref{eq:graph.subtasks}, the number of landmark points
$N_\mathit{l}$, and inner dimensions $d, r$
in~\eqref{eq:multi.kernel.nodimred} are identified by
grid-search.

\subsection{Cherry Hills water network}

\begin{figure}[t!]
  \centering
  \subfloat[$\hat{\vect{U}}_1^{(*)}$ \label{fig:plot.U1}]
  {\includegraphics[width =
    .45\columnwidth]{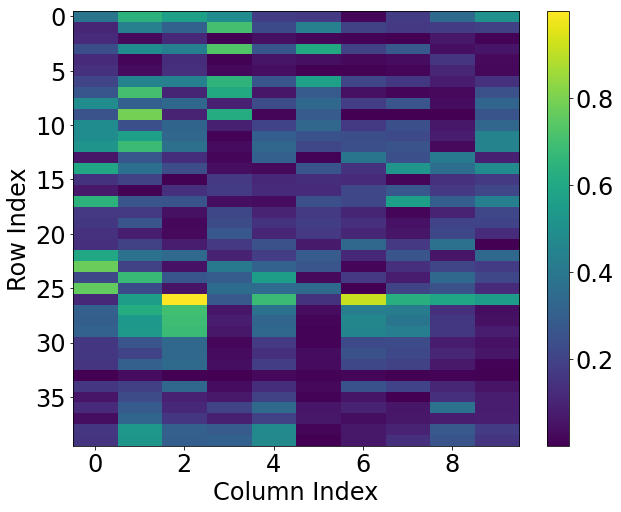}}
    \subfloat[$\hat{\vect{U}}_2^{(*)}$ \label{fig:plot.U2}]
  {\includegraphics[width = .44\columnwidth]{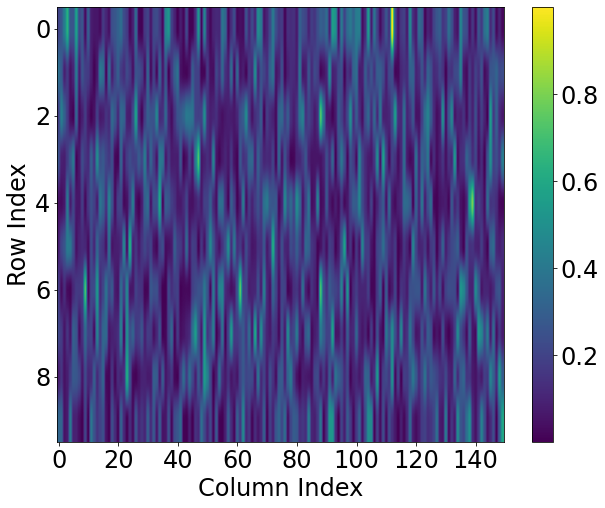}} \\
  \subfloat[$\hat{\vect{V}}_1^{(*)}$ \label{fig:plot.V1}]
  {\includegraphics[width =
    .45\columnwidth]{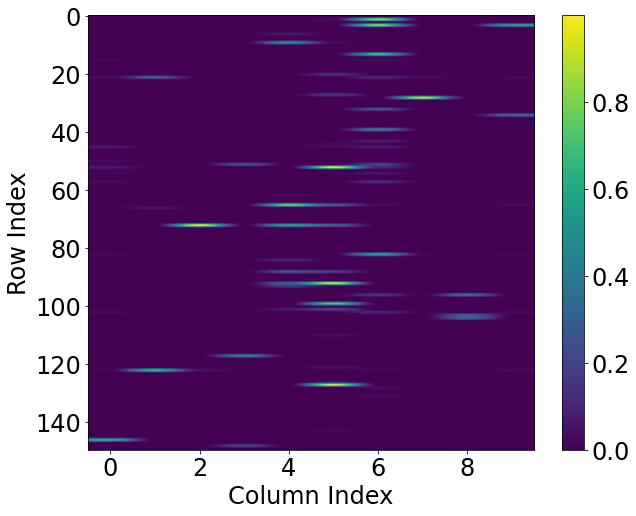}}
    \subfloat[$\hat{\vect{V}}_2^{(*)}$ \label{fig:plot.V2}]
  {\includegraphics[width = .43\columnwidth]{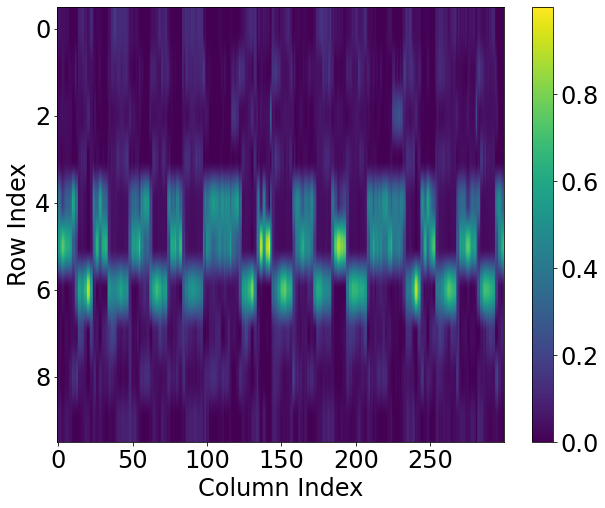}}
  \centering
  \caption{Visualization of coefficient matrices
    $\hat{\vect{U}}_1^{(*)}, \hat{\vect{U}}_2^{(*)},
    \hat{\vect{V}}_1^{(*)}$ and $\hat{\vect{V}}_2^{(*)}$,
    output by \cref{alg:multil.general} for Cherry Hills
    water network data, at sampling ratio $s = 0.5$. Heat maps
    show the entry-wise absolute values of the matrices,
    rescaled to be in $[0, 1]$ range.
  } \label{fig:plot.coeff}
\end{figure}

\begin{table}
  \centering
  \begin{tabular}{c||c|c|c}
    \hline
    Datasets $\setminus$ Methods & S-VAR & MMF & MultiL-KRIM \\
    \hline \hline
    Cherry Hills & \num{30000} & \num{6400} & \num{6400} \\
    Sioux Falls & \num{15000} & \num{2190} & \num{2190} \\
    \hline
  \end{tabular}
  \caption{Number of unknown parameters.}
  \label{tab:params}
\end{table}

Cherry Hills water network consists of \num{36} nodes
(0-simplices), \num{40} pipes (1-simplices), and \num{2}
triangles
(2-simplices)~\cite{leblanc1975algorithm}. Time-varying
water flow over the network is generated using the EPANET
software~\cite{rossman2010overview}. Flows are measured
hourly in cubic meter per hour (m$^3$/h) in a span of
\num{300} hours. The size of data $\vect{Y}$ is
$40\times 300$, and the number of landmark points is
$N_\mathit{l}=150$.  \cref{fig:plot.water} shows MAE values
of competing methods across different sampling
ratios. MultiL-KRIM outperforms the competing methods at
every sampling ratio. It significantly outperforms
S-VAR~\cite{krishnan2023simplicial, krishnan2024simplicial,
  money2024evolution}, especially at higher sampling ratios.
FlowSSL~\cite{jia2019graph}, which solely depends on the
Hodge Laplacians-based regularizers, scores much higher MAE
values than the other methods. MultiL-KRIM outperforms also
the blind matrix-factorization
MMF~\cite{cichocki2007multilayer}.

\cref{fig:plot.coeff} visualizes the coefficient matrices
learned by \cref{alg:multil.general}. In particular,
\cref{fig:plot.V1,fig:plot.V2} draw the heatmaps of
$\hat{\vect{V}}_1^{(*)}$ and $\hat{\vect{V}}_2^{(*)}$ of
\cref{alg:multil.general}, respectively.  The heatmaps show
that $\hat{\vect{V}}_2^{(*)}$ and especially
$\hat{\vect{V}}_1^{(*)}$ are sparse. The portion of entries
whose absolute value is larger than \num{0.001} is
\num{9.1}\% in $\hat{\vect{V}}_1^{(*)}$ and \num{89.6}\% in
$\hat{\vect{V}}_2^{(*)}$. In contrast, by
\cref{fig:plot.U1,fig:plot.U2},
$\hat{\vect{U}}_1^{(*)}, \hat{\vect{U}}_2^{(*)}$ are dense,
where the percentage of entries whose absolute value is
larger than \num{0.001} is \num{98.8}\% in
$\hat{\vect{U}}_1^{(*)}$ and \num{99.8}\% in
$\hat{\vect{U}}_2^{(*)}$.

\subsection{Sioux Falls transportation network}

The Sioux Falls transportation network has \num{24} nodes
(0-simplices), \num{38} edges (1-simplices), and \num{2}
triangles (2-simplices) \cite{rossman1994modeling}. The
synthetic time-varying traffic flow is generated as in
\cite{krishnan2023simplicial,money2024evolution}, so that it
contains both divergence flows and cyclic flows.  The
measurement unit is neglected, simply denoted as ``units.''
The size of data $\vect{Y}$ is $38\times 300$, while
$N_\mathit{l} = 50$. \cref{fig:plot.traffic} plots MAE
curves against sampling ratios. MultiL-KRIM continues to
outperform the competing methods, with notable gaps when
compared with S-VAR~\cite{krishnan2023simplicial,
  krishnan2024simplicial, money2024evolution} and
FlowSSL~\cite{jia2019graph}. Meanwhile, the gap between
MultiL-KRIM and MMF~\cite{cichocki2007multilayer}
accentuates at lower sampling
ratios. FlowSSL~\cite{jia2019graph} scores again the highest
MAE values.

With regards to dimensionality reduction, \cref{tab:params}
compares the number of unknown parameters in
S-VAR~\cite{krishnan2023simplicial, krishnan2024simplicial,
  money2024evolution}, MMF~\cite{cichocki2007multilayer},
and MultiL-KRIM. Although the number of parameters of
MultiL-KRIM are \num{21.3}\% (Cherry Hills) and \num{14.6}\%
(Sioux Falls) of those of S-VAR, MultiL-KRIM still
outperforms in both datasets.


\section{Conclusions}

This paper applied the recently developed MultiL-KRIM, a
multilinear kernel-regression-imputation and
manifold-learning framework, into the time-varying edge-flow
imputation problem. MultiL-KRIM approximated data via simple
geometric arguments and facilitated computation by low-rank
matrix factorization, without the need for training data. To
further realize the common priors of edge flows,
MultiL-KRIM's inverse problem incorporated also the graph's
Hodge Laplacians. Numerical tests on real water and traffic
networks showed that MultiL-KRIM offers better performance than
several state-of-the-art methods.


\clearpage
\balance
\printbibliography[title = {\normalsize\uppercase{References}}]

\end{document}